\def\year{2022}\relax
\def\showauthors@on{T}
\documentclass[letterpaper]{article} 
\usepackage[submission]{aaai24}  
\usepackage{times}  
\usepackage{helvet}  
\usepackage{courier}  
\usepackage[hyphens]{url}  
\usepackage{graphicx} 
\usepackage{natbib}  
\usepackage{caption} 
\DeclareCaptionStyle{ruled}{labelfont=normalfont,labelsep=colon,strut=off} 
\usepackage{algorithm}
\usepackage{algorithmic}

\usepackage{newfloat}
\usepackage{listings}
\floatstyle{ruled}
\newfloat{listing}{tb}{lst}{}
\floatname{listing}{Listing}
\usepackage[utf8]{inputenc} 
\usepackage[T1]{fontenc}    
\usepackage{url}            
\usepackage{booktabs}       
\usepackage{amsfonts}       
\usepackage{nicefrac}       
\usepackage{microtype}      
\usepackage{amssymb}
\usepackage{multirow}
\usepackage{subcaption}
\PassOptionsToPackage{usenames,dvipsnames,table}{xcolor}
\usepackage[most]{tcolorbox}

\usepackage{amsthm}
\theoremstyle{definition}

\newtcolorbox{mybox}[1][]{
  colback=blue!5!white,
  colframe=blue!75!black,
  fonttitle=\bfseries,
  title=#1
}

\usepackage{ifthen}
\usepackage{amsmath}
\usepackage{lipsum}
\usepackage{fancyhdr}       
\usepackage{graphicx}       
\usepackage[ruled,algo2e]{algorithm2e}

\usepackage{booktabs}
\usepackage{pifont}
\usepackage{listings}
\usepackage{graphicx}
\usepackage{subcaption}
\usepackage{breqn}
\usepackage{pgfplots}
\pgfplotsset{compat=1.16}
\usetikzlibrary{decorations.markings}
\usetikzlibrary{calc}

\title{Steering Language Generation: Harnessing Contrastive Expert Guidance and Negative Prompting for Coherent and Diverse Synthetic Data Generation}

\author{
    Charles O'Neill,\textsuperscript{\rm 1}
    Yuan-Sen Ting, \textsuperscript{\rm 2,3,4}
    Ioana Ciuc\u{a}, \textsuperscript{\rm 2,3}
    Jack Miller, \textsuperscript{\rm 1}
    Thang Bui \textsuperscript{\rm 1}
}
\affiliations{
    \textsuperscript{\rm 1}\textit{Mathematical Sciences Institute}, Australian National University\\
    \textsuperscript{\rm 2}\textit{Research School of Astronomy \& Astrophysics}, Australian National University\\
    \textsuperscript{\rm 3}\textit{School of Computing}, Australian National University\\
    \textsuperscript{\rm 4}\textit{Department of Astronomy}, The Ohio State University\\
}

\usepackage{bibentry}

\begin{document}

\maketitle

\begin{abstract}
Large Language Models (LLMs) hold immense potential to generate synthetic data of high quality and utility, which has numerous applications from downstream model training to practical data utilisation. However, contemporary models, despite their impressive capacities, consistently struggle to produce both coherent and diverse data. To address the coherency issue, we introduce contrastive expert guidance, where the difference between the logit distributions of fine-tuned and base language models is emphasised to ensure domain adherence. In order to ensure diversity, we utilise existing real and synthetic examples as negative prompts to the model. We deem this dual-pronged approach to logit reshaping as STEER: Semantic Text Enhancement via Embedding Repositioning. STEER operates at inference-time and systematically guides the LLMs to strike a balance between adherence to the data distribution (ensuring semantic fidelity) and deviation from prior synthetic examples or existing real datasets (ensuring diversity and authenticity). This delicate balancing act is achieved by dynamically moving towards or away from chosen representations in the latent space. STEER demonstrates improved performance over previous synthetic data generation techniques, exhibiting better balance between data diversity and coherency across three distinct tasks: hypothesis generation, toxic and non-toxic comment generation, and commonsense reasoning task generation. We demonstrate how STEER allows for fine-tuned control over the diversity-coherency trade-off via its hyperparameters, highlighting its versatility.
\end{abstract}

\section{Introduction}
Large language models (LLMs) learn a compressed representation of human knowledge through understanding language constructs, grammar and semantics as conditional probability distributions over subword tokens \cite{wei2022emergent, shanahan2023talking, zhao2023survey}. The prowess of LLMs in autoregressively generating text indicates a profound internalisation of our world view expressed through language \cite{gilbert2023semantic}. This internalised knowledge brings the potential of LLMs to not only parse, understand, and respond to data, but also to synthesise such data autonomously \cite{veselovsky_generating_2023, halterman2023synthetically, platzer_rule-adhering_2022, tan2022using}. Despite their prowess, the generation of synthetic data by these models presents a significant challenge, specifically in terms of data coherence and diversity \cite{alaa_how_2022, lu2023machine}. Specifically, there seems to be a trade-off between two measures of data quality: (1) how well synthetic samples resemble real samples (\textbf{fidelity}); and (2) how broadly the synthetic samples cover the real distribution (\textbf{diversity}); see Figure \ref{fig:tradeoff}. The current state-of-the-art often struggles to generate data that maintains semantic fidelity, embraces diversity, and transcends mere reproduction of the training set \cite{li2022faithfulness}. Whilst larger models and optimised training may eventually mitigate this somewhat, a more intriguing question remains: \textit{how can we reshape the probabilistic token selection at inference time in order to achieve better synthetic data generation?}


To this end, we address the challenge of synthetic data generation with \textbf{S}emantic \textbf{T}ext \textbf{E}nhancement via \textbf{E}mbedding \textbf{R}epositioning (\textbf{STEER}). The main contribution of STEER is \textit{contrastive expert guidance}, a logit reshaping technique that emphasises the distinctions between a fine-tuned (domain) model and the base model, attracting the generator towards producing text characteristic of the specific domain. Simultaneously, it utilises \textit{negative prompting}, another logit modification approach, which discourages the production of tokens found in previously generated synthetic or real examples, thereby ensuring the generation of synthetic examples that are both novel and representative. STEER operates at inference time and is architecture agnostic, using fine-tuning to delineate the semantic space of interest.

Our proposed STEER method offers several key contributions:
\begin{enumerate}
\item We propose STEER, a strategy for generating synthetic data that navigates the trade-off between coherency and diversity. STEER relies on the controllability of conditional generation induced by contrastive expert guidance and negative prompting.
\item We evaluate STEER with metrics that capture the diversity and coherency of generate data. Our approach demonstrates superior performance over current decoding methods on three tasks: scientific hypothesis generation, toxic social media comment classification, and commonsense question-answering. Further, we validate the efficacy of STEER by training models on the synthetic data for downstream tasks such as classification, and assessing human preference of STEER data over data generated with other methods.
\item Finally, we provide ablation studies determining how STEER enables better management of the trade-off between coherency and diversity by tweaking the hyperparameters that control the contrastive expert guidance and negative prompting, respectively.
\end{enumerate}

\begin{figure}[htpb]
\vspace{-5pt}
\centering
\resizebox{1\columnwidth}{!}{%
\begin{tikzpicture}
    \tikzset{
        my arrow/.style={
            postaction=decorate,
            decoration={
                markings,
                mark=at position #1 with {\arrow{>}}
            }
        }
    }
    
    \begin{axis}[
        title={\textbf{A:} Contrastive expert guidance},  
        no markers, 
        domain=0:10, 
        samples=100,
        ymin=0,
        axis lines*=left, 
        xlabel=$x$,
        every axis y label/.style={at=(current axis.above origin),anchor=south},
        every axis x label/.style={at=(current axis.right of origin),anchor=west},
        height=4cm, 
        width=8cm,
        xtick=\empty, 
        ytick=\empty,
        enlargelimits=false, 
        clip=false, 
        axis on top,
        grid = major,
        axis lines = middle,
        y axis line style={draw=none},
        name=plot1
    ]

    \addplot [draw=none, fill=blue!20, opacity=0.6, domain=0:10] 
    {0.1*exp(-(x-2)^2 / (2*1)) + 0.15*exp(-(x-5)^2 / (2*1.3))} \closedcycle;

    \addplot [draw=none, fill=green!20, opacity=0.6, domain=0:10]
    {0.3*exp(-(x-5)^2 / (2*0.3))} \closedcycle;

    \addplot [dotted, ultra thick, fill=none, domain=3:7]
    {0.3*exp(-(x-5)^2 / (2*0.3))};

    \addplot [draw=none, fill=red!20, opacity=0.6, domain=0:10]
    {0.2*exp(-(x-5)^2 / (2*1.4))} \closedcycle;

    \addplot [my arrow=1.0, ultra thick, fill=none, domain=4.5:4.7]
    {0.3*exp(-(x-5)^2 / (2*0.3))};
    \addplot [my arrow=1.0, ultra thick, fill=none, domain=5.5:5.3]
    {0.3*exp(-(x-5)^2 / (2*0.3))};

    \end{axis}

    \node at (4.15,0.5) {$\mathrm{P}_\theta$};
    \node at (1.5,0.5) {$\mathrm{P}_\phi$};
    \node at (4.7,2) {$\widetilde{\mathrm{P}_\theta}(x_i\mid x_{j<i})$};

\begin{scope}[yshift=-3.8cm]  
\begin{axis}[  
    title={\textbf{B:} Negative prompting},  
    no markers, 
    domain=0:10, 
    samples=100,
    ymin=0,
    axis lines*=left, 
    xlabel=$x$,
    every axis y label/.style={at=(current axis.above origin),anchor=south},
    every axis x label/.style={at=(current axis.right of origin),anchor=west},
    height=4cm, 
    width=8cm,
    xtick=\empty, 
    ytick=\empty,
    enlargelimits=false, 
    clip=false, 
    axis on top,
    grid = major,
    axis lines = middle,
    y axis line style={draw=none},
    name=plot2
]
    \addplot [draw=none, fill=red!20, opacity=0.5, domain=0:10]
    {0.2*exp(-(x-5)^2 / (2*0.8))} \closedcycle;

    \addplot [draw=none, fill=yellow!20, opacity=0.75, domain=0:10]
    {0.15*exp(-(x-5)^2 / (2*1.2))} \closedcycle;

    \addplot [dotted, ultra thick, fill=none, domain=1:9]
    {0.15*exp(-(x-5)^2 / (2*1.2))};

    \addplot [my arrow=1.0, ultra thick, fill=none, domain=4:3.5]
    {0.15*exp(-(x-5)^2 / (2*1.2))};
    \addplot [my arrow=1.0, ultra thick, fill=none, domain=6:6.5]
    {0.15*exp(-(x-5)^2 / (2*1.2))};

\end{axis}
    \node at (3.22, 2.1) {$\mathrm{P}_\theta$};
    \node at (3.22,0.3) {$\widehat{\mathrm{P}_\theta}(x_i\mid x_{j<i})$};
\end{scope}  

\begin{axis}[
        name=smallPlot,
        title={\textbf{C:} Final sampling distribution},
        at={(7cm, -0.97cm)},  
        height=2.5cm,
        width=5cm,
        no markers, 
        domain=0:10, 
        samples=100,
        ymin=0,
        axis lines*=left, 
        xlabel=$x$,
        every axis y label/.style={at=(current axis.above origin),anchor=south},
        every axis x label/.style={at=(current axis.right of origin),anchor=west},
        xtick=\empty, 
        ytick=\empty,
        enlargelimits=false, 
        clip=false, 
        axis on top,
        grid = major,
        axis lines = middle,
        y axis line style={draw=none}  
    ]

    \addplot [draw=magenta, ultra thick, fill=magenta!20, opacity=0.5, domain=0:10]
    {0.2*exp(-(x-5)^2 / (2*1.2))} \closedcycle;

\end{axis}
\node at (8.7, -0.5) {$\overline{\mathrm{P}_\theta}$};

    \node[circle, draw=blue, fill=blue!20, opacity=0.3, ultra thick, minimum size=2cm] (circle) at (14cm,-0.5cm) {};

    \draw[->, thick, shorten >=5mm] ($(plot1.east)+(0,0.8cm)$) -- ($(smallPlot.north)+(0,0.7cm)$);
    \draw[->, thick, shorten >=5mm] (plot2.east) -- (smallPlot.south);
    
    \draw[->, thick, shorten >=5mm, shorten <=5mm] (smallPlot.east) -- (circle) node[midway, above] {\textbf{D:} Generate};

\foreach \i in {1,2,...,50} { 
    \pgfmathsetmacro{\angle}{rnd*360}
    \pgfmathsetmacro{\radius}{rnd*0.03cm}
    \node[circle, fill=blue, inner sep=1pt] at ($(circle.center)+(\angle:\radius)$) {};
}

\foreach \x in {-1,-0.8,...,1} {
    \foreach \y in {-1,-0.8,...,1} {
        \pgfmathsetmacro{\distance}{sqrt(\x*\x + \y*\y)}
        \ifdim \distance pt < 0.03cm
            \node[circle, fill=magenta, inner sep=1pt] at ($(circle.center)+(\x cm,\y cm)$) {};
        \fi
    }
}

\end{tikzpicture}%
}
\caption{\textbf{Synthetic data generation with STEER}. We first use a real dataset $\mathcal{D}r$ to fine-tune a generative language model $\mathrm{P}_\theta$. \textbf{A:} Contrastive expert guidance modifies the sampling distribution $\widetilde{\mathrm{P}}_\theta$ (green). \textbf{B:} Negative prompting downweights tokens in real/generated examples, leading to $\widehat{\mathrm{P}}_\theta$ (yellow). \textbf{C:} Final sampling distribution combines contrastive guidance and negative prompting. \textbf{D:} Sampling, with each generated example fed into negative prompts, creates a coherent and diverse synthetic dataset (magenta dots) closely resembling the real distribution (blue circle).}
\label{fig:main}
\end{figure}
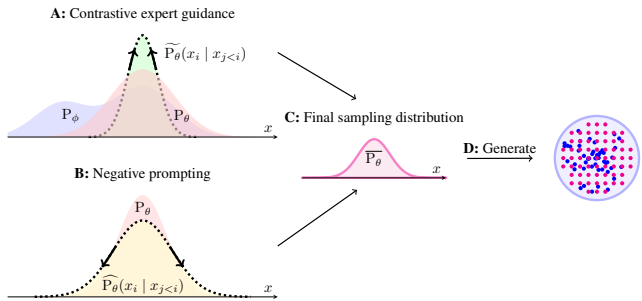

\section{Related work}

\subsection{Evolution of Generative Models and Control Over Conditional Generation}
Recent years have witnessed a remarkable advancement in generative artificial intelligence models, which typically ingest context, frequently presented as prompts, and subsequently generate text, images, videos, or audio conditioned upon this provided context \cite{jo2023promise, Epstein_2023}. The extent to which a model attends to such context is determined during the training phase, thereby granting users minimal control \cite{chung2022scaling}. This issue first garnered attention with the emergence of generative adversarial models (GANs) and then diffusion models \cite{dhariwal2021diffusion}. A series of techniques and methodologies have been proposed to better control this conditioning, forming the basis of the following sections.

Traditional efforts to guide language models, such as PPLM \cite{Dathathri2020Plug} and GeDI \cite{krause2020gedi}, relied on external classifiers for specific attributes \cite{sitdikov_classifiers_2022}. These methods utilise the gradient of a classifier, trained to recognise certain attributes, to adjust the logits during the generation process. By computing the gradient with respect to the desired attribute and applying it to the model's hidden states, they steer the generative process towards or away from particular characteristics. This proved expensive and complex, requiring additional models and continuous adjustments during implementation. 

Logit guidance emerged as an alternative approach for governing aspects of text generation, focusing on the manipulation of token distributions rather than semantic control, thereby signaling a progression towards architecture-agnostic guidance without additional classifiers. In the context of autoregressive language models, the token generation process unfolds sequentially, where the probability of each token is conditioned on its preceding tokens. Mathematically, the joint probability of a token sequence $w$ is expressed as:
\begin{equation}
\mathrm{P}_{\theta}(w) = \prod\limits_{i}^{N}\mathrm{P}_\theta(w_{i}|w_{j<i}).
\end{equation}
Here, $\mathrm{P}_\theta(w_i|w_{j<i})$ denotes the conditional probability of the $i$-th token given its predecessors, and is modeled as a distribution over the entire vocabulary. This distribution, represented as logits, is a function of the model's current state and can be directly manipulated. By selectively adjusting the logits' values, it's possible to exert influence over the generated text, steering it towards specific characteristics or themes. Logit guidance, therefore, leverages this property to offer fine-grained control over the generation process, fulfilling targeted objectives without the need for external models or classifiers.

For instance, one challenge we face is preserving the fidelity of synthetic examples, meaning the generated content should reflect the intended domain. Contrastive Decoding (CD) \cite{li_contrastive_2022} sought to overcome this challenge by enhancing the distinction between the log-probabilities of a sophisticated, high-capacity language model (LM) and a less capable, smaller counterpart. The log-probabilities of the tokens from the smaller model, $\mathrm{P}_{\phi}(w_i | w_{j<i})$, are subtracted from those of a larger, ``expert'' $\mathrm{P}_{\theta}(w_i | w_{j<i})$ model as a form of regularisation.
This is similar to the DExpert approach undertaken by \citet{liu-etal-2021-dexperts}, who perform interpolation in the output space as a form of ensembling. Intuitively, under the ensemble, tokens only get high probability if they are considered likely by the experts and unlikely by the anti-experts.

Drawing parallels to other logit adjustment methods, Context-Aware Decoding (CAD) also recalibrates token probabilities \cite{shi_trusting_2023}. CAD achieves this by dividing the log-probability of the upcoming token, when considering the model with input context, by the log-probability of the same token under the model sans context—effectively adjusting its contextual awareness. Advancing on this premise, \citet{malkin2022coherence} introduced a coherence-boosting (CB) strategy, differentially balancing the contributions of both full and partial contexts. This method illuminates early tokens by assessing the log difference between token distributions—with and without the inclusion of this early context. In doing so, the distribution hones in on tokens exhibiting high probability in the presence of early context but low probability when devoid of it. Similarly, \citet{su2022contrastive} used a degeneration penalty, defined as the maximum cosine similarity between the representation of a continuation and that of all previous tokens, to prevent model degeneration. They refer to this method as contrastive search.

Building upon the innovative efforts to exert control over generative models, the concept of classifier-free guidance (CFG) emerged as a further refinement to the field. It embodies a methodology aligning closely with previous advancements but distinctively eliminating the reliance on separate classifier models. The genesis of CFG can be traced back to diffusion models, where techniques were developed to reshape the latent sampling distribution to synchronise more harmoniously with the prompt \cite{dhariwal_diffusion_2021}. Addressing earlier complexities, \citet{ho_classifier-free_2022} synthesised the classifier's role into the model's training process itself, paving a new path for efficiency.

In the context of autoregressive language models, which inherently excel in unconditional generation, CFG represents a natural evolution \cite{sanchez_stay_2023}. By manipulating the generation of subsequent tokens to accentuate the conditioning on the prompt, it builds upon the existing framework of logit guidance and log-probability adjustment. This manipulation can be formally expressed as follows:
{\scriptsize
\begin{align}
\label{eq:cfg_logit}
\log \widehat{\mathrm{P}_\theta}\left(w_i \mid w_{j<i}, c\right)&=\log \mathrm{P}_\theta\left(w_i \mid w_{j<i}\right) 
+\gamma\Big(\log \mathrm{P}_\theta\left(w_i \mid w_{i<j}, c\right)\nonumber \\
&-\log \mathrm{P}_\theta\left(w_i \mid w_{j<i}\right)\Big).
\end{align}
}
CFG also allows for the avoidance of specific aspects of generation through the use of negative prompting. By adjusting $\gamma$ to be negative in the equation above, control is exerted to guide generation away from a given prompt. This methodology has found exceptional efficacy in diffusion models \cite{Andrew_2023, crowson2022vqganclip, du2020compositional, rombach2022highresolution, miyake2023negative}, further enriching the spectrum of control over generative processes. Both CFG and negative prompting exploit the latent semantic information in token predictions \cite{winterhalder2021latent, jiang2023latent, lee2019mathematical, durrani2022transformation}.

\subsection{Synthetic data generation leveraging language models}

In the time before today's large language models (LLMs), traditional augmentation of text data was practiced through techniques like random text edits, synonym replacements, masked word predictions, reinforcement learning, lossy translation, and text blending \cite{feng_survey_2021, wei-zou-2019-eda, zhang2016characterlevel, ng-etal-2020-ssmba, liu-etal-2020-data, fadaee-etal-2017-data, zhang2018mixup, guo-etal-2020-sequence, sun-etal-2020-mixup, zhang-etal-2022-treemix}. Further advancements involved fine-tuning of models and conditional generation implementation, but the discrete nature of language hindered traditional methods' effectiveness \cite{xia2020cgbert, yoo2018data, hou-etal-2018-sequence, wu2022text}. Challenges led to the utilisation of LLMs as synthetic data generators, using methods like fine-tuning conditional generation and zero-shot/few-shot prompting. This included training with prompts, fine-tuning with specific text structures, and leveraging both techniques with additional real data \cite{kumar_data_2021, yoo-etal-2021-gpt3mix-leveraging, sahu-etal-2022-data, yuan2021synthbio, hartvigsen-etal-2022-toxigen, moller_is_2023, eldan_tinystories_2023, josifoski_exploiting_2023, he_generate_2022, veselovsky_generating_2023}.

Many studies acknowledge a distribution difference between real and synthetic examples, reflecting poor fidelity. The generation of low fidelity data often requires filtration, using classifiers to weed out synthetic instances, or costly human evaluation \cite{veselovsky_generating_2023, chung_increasing_2023}. As for diversity, existing techniques like logit suppression and temperature raising struggle with a diversity-fidelity tradeoff, leading to low-quality generations that need human quality control \cite{chung_increasing_2023}. Some strategies leverage taxonomy-based generation or multiple inference runs with different seeds to encourage diversity, although at increased computational costs \cite{veselovsky_generating_2023, josifoski_exploiting_2023}. As fine-tuning existing models for the purpose of synthetic text generation becomes easier and more efficient, it's clear that more controllable and rigorous methods are required in order to navigate the trade-off between fidelity and diversity.

\subsection{Evaluation frameworks for synthetic textual data}

The evaluation of the quality of generated data has traditionally been simplistic, often relying on downstream task performance or employing statistical divergence measures like Kullback-Leibler (KL) divergence and Fréchet distance, as well as coherency measures like MAUVE and cosine similarity \cite{veselovsky_generating_2023, platzer_rule-adhering_2022, sitdikov_classifiers_2022, josifoski_exploiting_2023, he_generate_2022, moller_is_2023, heusel2018gans, pillutla2021mauve, li_contrastive_2022}. These approaches are limited in their ability to quantify fidelity and diversity comprehensively, are often inapplicable to certain models like GANs due to inaccessible likelihoods or suffer from problems in high-dimensional spaces \cite{wu2017on, goodfellow2014generative, kingma2022autoencoding, theis2016note}. Previous precision-recall analyses have shown shortcomings like mode collapse, lack of robustness, and inability to evaluate authenticity \cite{sajjadi2018assessing, flach2015precision, naeem2020reliable, adlam2019investigating, meehan2020nonparametric}. Attempts have been made to propose improved metrics, but no evaluation schema has emerged as the gold-standard, emphasising the need for nuanced, sample-level evaluation to effectively triage quality and automate culling of low-quality instances \cite{alaa_how_2022}. As such, the evaluation of synthetic data is typically conducted by employing a wide range of metrics for each dimension of quality we wish to assess.

\section{STEER}
\label{sec:steer}

We denote a real datapoint as $X_r$ from the real distribution $\mathbb{P}_r$ and a synthetic example as $X_s$ from the learned synthetic distribution $\mathbb{P}_s$. This then induces the definitions of the real dataset $\mathcal{D}_r=\{X_{r,i}\}^n_{i=1}$ with $n$ distinct data points and $\mathcal{D}_s = \{X_{s, j}\}_{j=1}^m$ with $m$ distinct data points. 
In order to use both contrastive logit reshaping and negative prompting, we must first have the ability to unconditionally sample tokens from the distribution of interest i.e. the distribution of the real data, $\mathbb{P}_r$. We leverage transfer learning by starting with a pre-trained transformer architecture. The fine-tuning process employs a next-word prediction paradigm, wherein the model is conditioned on prompts such as 'Generate a scientific hypothesis:' and asked to autoregressively predict the next word in real examples. This provides us with a model $\mathrm{P}_\theta$ that is proficient at generating semantically similar examples in response to the given prompt, effectively sampling from the distribution $\mathbb{P}_r$. We denote $\mathrm{P}_\theta$ as the \textit{domain model}. We also retain the model before fine-tuning as the \textit{base model} $\mathrm{P}_\phi$.

\paragraph{Contrastive expert guidance for fidelity}
STEER first seeks to reweight the importance of the target distribution itself in order to generate high-fidelity text that is from said distribution. The contrastive objective $\widetilde{\mathrm{P}_{\theta}}$ seeks to maximise the likelihood of the domain model's sequence, while minimising the likelihood of the same sequence under the base model's distribution.

\begin{mybox}[Contrastive Expert Guidance]
Let \(\mathrm{P}_\theta\) be the domain model capable of sampling from \(\mathbb{P}_r\), and \(\mathrm{P}_\phi\) be the base model. The modified logit sampling distribution we denote as \(\widetilde{\mathrm{P}_{\theta}}\), which is defined as:
{\small
\begin{align*}
\label{eq:ceg}
\log \widetilde{\mathrm{P}_{\theta}}(w_{i} | w_{j<i}) &= \log \mathrm{P}_\theta(w_i | w_{j<i}) - \gamma \log \mathrm{P}_\phi(w_i | w_{j<i})
\end{align*}}where $\gamma \in [0,1]$ is a hyperparameter that controls the strength of the contrastive expert guidance.
\end{mybox}

In our approach, the fine-tuned domain model $\mathrm{P}_\theta$ is sensitive to the specifics of the target domain, contrasting with the base model $\mathrm{P}_\phi$, trained on a general language task. The contrastive objective leverages this difference to steer the generation towards text that aligns more with the domain distribution $\mathbb{P}_r$ than with the broader distribution of the base model, the emphasis on which is controlled by hyperparameter $\gamma$. 

Consequently, our synthetic examples, generated via contrastive guidance, are not only plausible—reflecting high probability under the expert language model—but also distinctive to the target domain—exhibiting lower probability under the less-specialised language model. This balance guides synthetic example generation towards text that faithfully represents the target domain distribution $\mathcal{D}_r$ while still exploring the semantic space comprehensively.

\paragraph{Negative prompting for diverse and authentic generation}
To encourage diversity and originality in synthetic examples, we complement our contrastive objective with a negative prompting mechanism. By integrating a negative prompt $\bar{c}$—comprising tokens from previously generated synthetic examples and real examples from $\mathcal{D}_r$—we steer the model towards novel sequence generation. This is achieved by creating another logit distribution $\widehat{\mathrm{P}_\theta}$:

\begin{mybox}[Negative Prompting]
Let \(\mathrm{P}_\theta\) be the domain model and \(\bar{c}\) be the negative prompt. The logit distribution \(\widehat{\mathrm{P}_\theta}\) is defined as:
{\scriptsize
\begin{align*}
\log \widehat{\mathrm{P}_\theta}(w_i|w_{j<i},\bar{c}) &= \log \mathrm{P}_\theta(w_i|w_{j<i},\bar{c})\\
&+ \eta \Big(\log \mathrm{P}_\theta(w_i| w_{j<i}) - \log \mathrm{P}_\theta(w_i|w_{j<i},\bar{c})\Big)
\end{align*}}where $\eta \in [0,1]$ is a hyperparameter that controls the strength of the negative prompting.
\end{mybox}

The negative prompt $\bar{c}$ adjusts the next token's log probability, diminishing the likelihood of tokens found in $\bar{c}$. The novelty level of the synthetic examples is regulated by the parameter $\eta$. In practice, we maintain a dynamic set of real and synthetic examples, with the tokens therein constituting our negative prompt $\bar{c}$. This strategy, while maintaining domain-specificity via contrastive decoding, promotes diversity and novelty by discouraging the generation of previously encountered or existing examples.

\paragraph{Final modified logits}
This combination of contrastive expert guidance with negative prompting ensures a fine balance between adhering to the domain distribution and maintaining diversity, resulting in the generation of synthetic examples that are both novel and representative of the real dataset $\mathcal{D}_r$. The final distribution used to sample the next token during decoding is obtained by modifying the logits (i.e., the inputs to the softmax function that calculates the probabilities of the next token) of the domain model according to the contrastive objective and the negative prompting. The modified logit distribution $\overline{P_\theta}$ is given by:

\begin{mybox}[STEER]
Let \(\widetilde{\mathrm{P}_{\theta}}\) be the contrastive expert guidance-modified logit distribution, and \(\widehat{\mathrm{P}_\theta}\) be the logit distribution with negative prompting. The modified logit distribution \(\overline{\mathrm{P}_\theta}\) is defined as:
{\small
\begin{equation*}
\label{eq:steer}
\log \overline{\mathrm{P}_\theta}(w_i | w_{j<i}) = \log \widetilde{\mathrm{P}_{\theta}}(w_{i} | w_{j<i}) + \log \widehat{\mathrm{P}_\theta}(w_i|w_{j<i},\bar{c}).
\end{equation*}
}
\end{mybox}


Once we have the modified logits, we can perform nucleus sampling to generate the next token in the sequence. We continue this process until we generate a complete synthetic example. Repeating this process, sampling negative prompts from the real and current synthetic examples each time, yields a synthetic text dataset. This process is outlined in Algorithm \ref{alg:steer} in the Appendix.



\section{Methodology}

\subsection{Datasets Description}

Three distinct datasets were chosen to validate the generality and flexibility of the proposed method. The \textit{Arxiv Hypotheses} dataset consists of 10,000 scientific hypotheses extracted from Arxiv astronomy abstracts (that is, abstracts with the \texttt{astro.ph} tag) using GPT-3.5. This provides a complex semantic space suitable for assessing generative model fidelity. The \textit{Jigsaw Toxic Comments} dataset from Kaggle comprises user comments labeled for toxicity and facilitates conditional data generation, enabling downstream classification. We selected 15,000 comments with a positive toxicity label and 15,000 comments with a negative toxicity label for a total of 30,000 examples. The third dataset, \textit{CommonsenseQA}, developed by the Allen Institute for AI, contains 12,247 multiple-choice questions and offers a challenging platform to evaluate the model's understanding of intricate semantics and general world knowledge. We extracted the questions, multiple-choice options and answers to form one string for each example. For more detail on data curation, see the Appendix.

\subsection{Evaluation metrics}
The evaluation process for our proposed method begins with the instruction fine-tuning of the Falcon-7B open-source model to perform generation using the three datasets. Specifically, the model is given the real dataset of examples as instruction-completion pairs. For instance, the astronomy hypothesis generation task had the instruction ``Generate a scientific hypothesis about astronomy''. These fine-tuned models are used to generate 1000 examples for each dataset using the same instruction as in fine-tuning, using greedy decoding, nucleus sampling \cite{holtzman2020curious}, contrastive search sampling \cite{li_contrastive_2022} and STEER sampling. STEER itself uses nucleus sampling with its additional logit reshaping. To ensure balance for downstream tasks and comparison, uniform frequencies of each label are generated for the \textit{Jigsaw Toxic Comments} and \textit{CommonsenseQA} datasets. For CommonsenseQA, we generated not only the multiple-choice question and options but also the answer at the end of the generation, in the form \texttt{<Question> A. <Option A> B. <Option B> ... E. <Option E>. Answer: <answer>}. In parallel, 1000 samples are randomly selected from a corresponding holdout test set of real dataset (with the same uniform distribution of labels for the latter two tasks). All comparative metrics presented below refer to these 1000 samples, both real and synthetic.

\subsubsection{Diversity}
The metrics used to assess diversity are \textit{normalised $n$-grams} (in our case we choose $n=3$ and \textit{diversity score}.

\textit{Normalised $n$-grams.} This is calculated as the proportion of duplicated $n$-grams in the text: $\text{norm-$n$} = 100 \times \left(1.0 - \frac{|\text{unique $n$-grams}|}{|\text{total $n$-grams}|}\right)$ \cite{welleck2019neural}. This metric is considered an indicator of linguistic richness, where a high value corresponds to a
greater variety in language use.

\textit{Diversity score.} This is a product of repetition measures at different $n$-gram levels, typically defined by $\prod_{n=2}^4 \left(1.0 -\frac{\text{norm-$n$}}{100}\right)$.

\subsubsection{Coherence}
We use three metrics to assess coherence: \textit{cosine similarity} between real and synthetic datasets, \textit{MAUVE}, and \textit{adversarial AUROC}. These measures rely on embeddings for each example (real and synthetic), which are calculated using OpenAI's \texttt{text-embedding-ada-002} model, which calculates an embedding of length 1536 for each sample.

\textit{Cosine similarity.} As a measure of semantic coherence between datasets, we calculate the cosine similarity between the mean real embeddings and the mean synthetic embeddings: $v_r^\top v_s/(||v_r|| \cdot ||v_s||)$. A larger cosine similarity suggests greater coherence with the real dataset.

\textit{MAUVE.} MAUVE calculates information divergences in a quantised embedding space and thus measures token-distribution similarity between real and synthetic data \cite{pillutla2021mauve}. MAUVE first quantises the embedding space into a finite number of bins, and then calculates the Kullback-Leibler divergence between the real and synthetic data distributions in each bin. The average of the Kullback-Leibler divergences across all bins is the MAUVE score. A higher MAUVE score means a closer similarity between distributions.

\textit{Adversarial AUROC.} We train an adversarial classifier to distinguish between real and synthetic data. The idea is that more coherent synthetic data will lead to a lower AUROC score for this classifier, as it becomes more difficult to distinguish between real and synthetic examples.

\subsubsection{Downstream performance}
For downstream performance evaluation, the type of validation is determined by the downstream task. For the \textit{Arxiv Hypotheses}, a set of eight expert annotators were shown a sequence of hypothesis pairs, one being an example generated with nucleus sampling and the other with STEER. They were asked to annotate which one they preferred in terms of creativity and plausibility, without knowing which was which. The performance recorded in Figure \ref{fig:heatmaps} refers to the proportion of times the STEER sample was deemed to be better than the nucleus sample. A one-sided $Z$-test for proportions was used, testing the null hypothesis that the win rate was above $0.5$. For the \textit{Jigsaw Toxic Comments} and \textit{CommonsenseQA} datasets we train a classifier and a question-answerer, respectively, on the synthetic data. The classifier for Jigsaw was an XGBoost model trained on the synthetic embeddings. The question-answering model for CommonsenseQA was a fine-tuned BERT model with a classification head. The aim is to demonstrate superior downstream performance of the STEER dataset.

\section{Results}

\begin{table*}[htbp]
  \centering
  \footnotesize
  \setlength{\abovecaptionskip}{10pt}
  \begin{tabular}{l|l|c|c|c|c|c}
    \toprule
    & & Normalised n-grams & Diversity & Cosine Similarity & MAUVE & Adversarial AUROC \\
    \midrule
    \multirow{4}{*}{\rotatebox[origin=c]{90}{\textit{ArXiv}}} 
    & Top-$k$ & 0.44 & 0.06 & 0.83 & 0.73 & \textbf{0.61} \\
    & Nucleus & 0.38 & 0.04 & 0.83 & 0.72 & 0.64 \\
    & Contrastive & 0.31 & 0.03 & 0.83 & 0.17 & 0.85 \\
    & STEER & \textbf{0.65} & \textbf{0.10} & \textbf{0.84} & \textbf{0.75} & 0.66 \\
    \midrule
    \multirow{4}{*}{\rotatebox[origin=c]{90}{\textit{Jigsaw}}}
    & Greedy & 0.55 & 0.12 & 0.70 & 0.11 & 0.99 \\
    & Nucleus & 0.65 & 0.21 & 0.71 & 0.14 & 0.99 \\
    & Contrastive & 0.61 & 0.16 & \textbf{0.73} & 0.08 & 0.99 \\
    & STEER & \textbf{0.73} & \textbf{0.28} & \textbf{0.73} & \textbf{0.30} & 0.99 \\
    \midrule
    \multirow{4}{*}{\rotatebox[origin=c]{90}{\textit{QA}}}
    & Greedy & 0.54 & 0.12 & 0.76 & 0.76 & 0.95 \\
    & Nucleus & 0.55 & 0.12 & 0.77 & 0.80 & 0.96 \\
    & Contrastive & 0.49 & 0.09 & 0.77 & 0.22 & 0.97 \\
    & STEER & \textbf{0.62} & \textbf{0.18} & \textbf{0.78} & \textbf{0.84} & \textbf{0.92} \\
    \bottomrule
  \end{tabular}
  \caption{Comparison of normalised n-grams, diversity, cosine similarity, MAUVE, and adversarial AUROC for a fine-tuned Falcon-7B across three datasets: ArXiv Hypotheses, Jigsaw Toxic Comments, and CommonsenseQA. Except for adversarial AUROC, higher is better. Here, ``Contrastive'' stands for ``Contrastive Search'' \cite{su2022contrastive}.}
  \label{tab:model_performance}
  \vspace{-10pt}
\end{table*}

Table \ref{tab:model_performance} presents a comparative study of STEER with alternative decoding and sampling methods, namely greedy decoding, nucleus sampling, and contrastive search sampling. We evaluate the synthetic datasets of 1000 samples generated using these decoding methods on the basis of normalised n-grams, diversity, cosine similarity, MAUVE, and adversarial AUROC for a fine-tuned Falcon-7B model.

\begin{figure}
  \centering
  \includegraphics[width=\columnwidth]{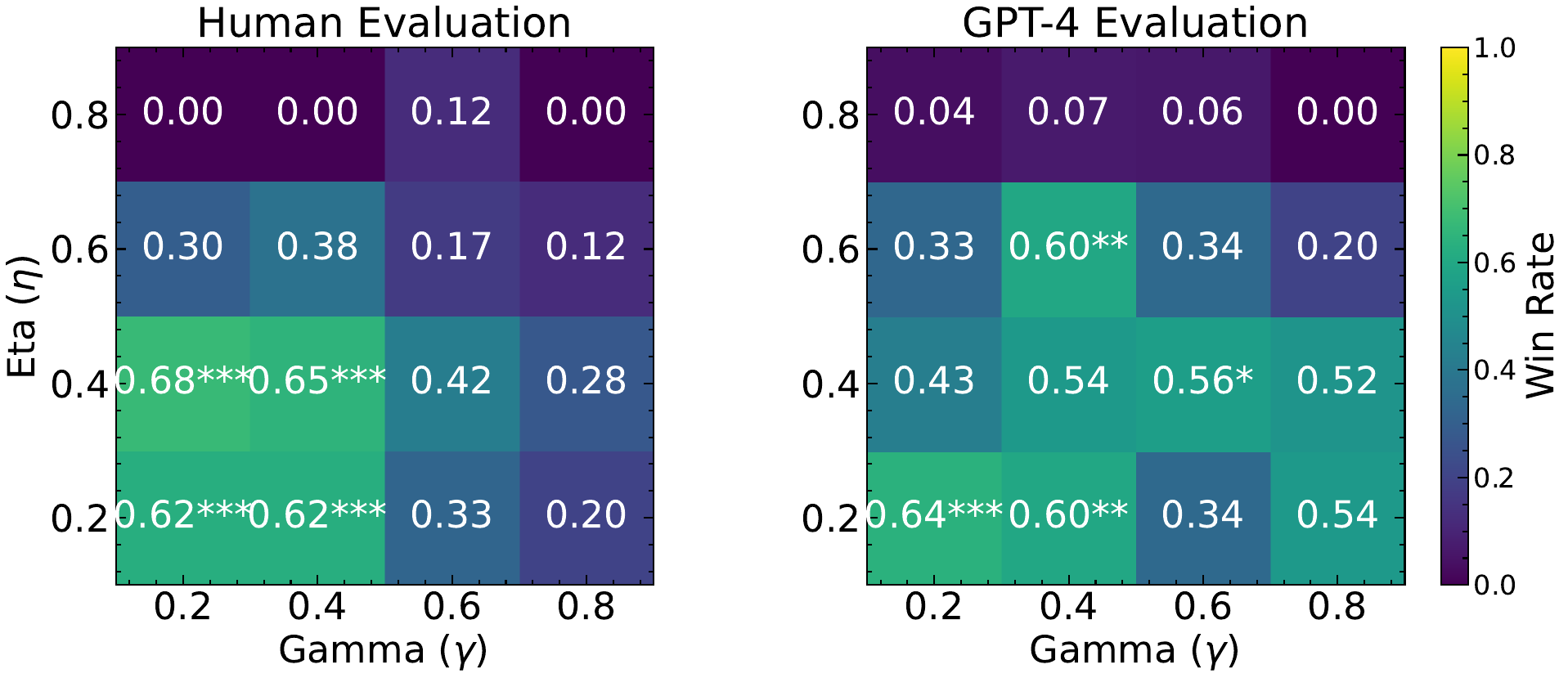}
  \caption{Win rate of STEER against nucleus sampling in the hypothesis generation task. The levels of significance are marked as follows: $***$ denotes $p<0.001$, $**$ denotes $0.001\leq p<0.01$, and $*$ denotes $0.01< p \leq 0.05$. Left denotes human evaluation, right is GPT-4.}
  \label{fig:combined_heatmap}
  \vspace{-10pt}
\end{figure}

\begin{figure}
\centering
\includegraphics[width=\columnwidth]{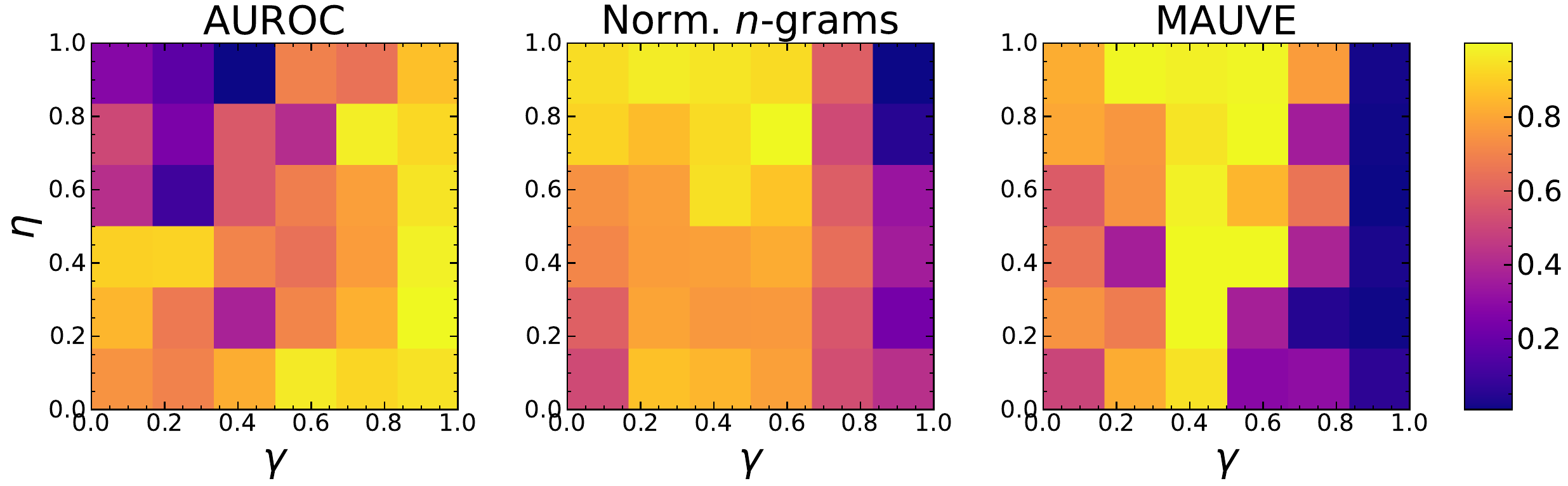}
\caption{Performance of Falcon-7B on the hypothesis generation task when varying the contrastive guidance hyperparameter $\gamma$ and the negative prompting hyperparameter $\eta$. 50 examples were produced for each combination of $\gamma$ and $\eta$ to evaluate the metrics on. A lower AUROC is better, and higher normalised $n$-grams and MAUVE are better.}
\label{fig:heatmaps}
\vspace{-10pt}
\end{figure}

The downstream performance comparison of STEER against other methods for four different models is encapsulated in Figure \ref{fig:heatmaps} and Table \ref{tab:downstream_performance}. These evaluations provide a comparative analysis of STEER's performance with real data and data generated with several alternative methods. STEER outperformed all methods in the \textit{Jigsaw} and \textit{CommonsenseQA} tasks for classification accuracy. Further, STEER had a significantly positive win rate over nucleus-generated samples in the \textit{Hypothesis} task for certain values of $\gamma$ and $\eta$, for both human and GPT-4 evaluators. We noticed that astronomy domain experts tended to have lower win-rates of STEER compared to general evaluators. Further comparisons between astronomy domain experts and general human evaluators are presented in the appendix.

Finally, we investigated the impact of the contrastive expert guidance hyperparameter $\gamma$ and the negative prompting hyperparameter $\eta$ on both diversity and coherence. Figure \ref{fig:heatmaps} demonstrates how key diversity and coherence metrics are affected as we vary each of the hyperparameters. 50 synthetic Arxiv hypothesis examples were generated for each combination of $\gamma$ and $\eta$, and the adversarial AUROC, normalised number of n-grams and MAUVE scores were calculated. In Figure \ref{fig:vary_one}, we provide a notion of how generation diversity (approximated by diversity score) and generation coherency (approximated by MAUVE) vary as we vary one hyperparameter at a time (including the number of negative prompts), keeping the other constant. Whilst this is limited by the context size of the model (2048 tokens for Falcon-7B), we use the maximum number of negative prompts that fit into the context size of the model when the prompt is also included.

\section{Discussion}

In this study, we offer a comprehensive analysis of STEER, examining its capabilities and performance across diverse model architectures and various downstream tasks. Our investigation not only sheds light on its potential advantages but also uncovers constraints that open new avenues for research in logit manipulation, with the ultimate goal of harnessing STEER for synthetic text generation across multiple domains and downstream tasks.

\begin{figure}
\centering
\includegraphics[width=\columnwidth]{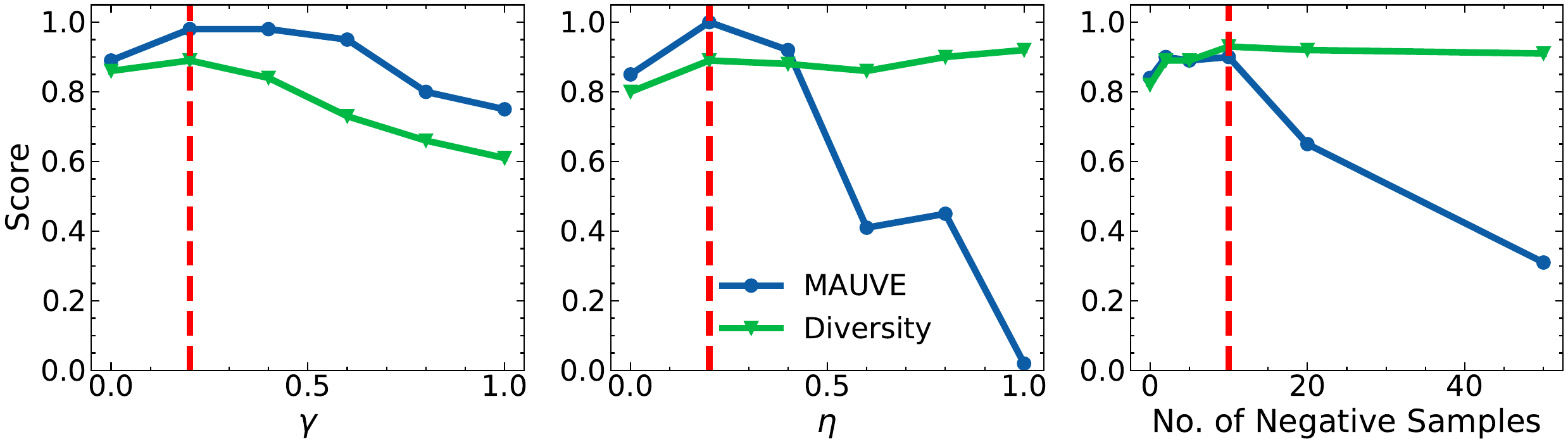}
\caption{Trade-offs when varying one hyperparameter at a time, keeping the other fixed at 0 (for $\gamma$ and $\eta$, which is not necessarily the optimal value). For the number of negative prompts, we set $(\gamma, \eta)=(0.4,0.4)$. The dashed red vertical line shows the point at which the sum of MAUVE and diversity score is greatest.}
\label{fig:vary_one}
  \vspace{-10pt}
\end{figure}

Table \ref{tab:model_performance} summarises the main results, demonstrating that STEER, even with minimal hyperparameter tuning, surpasses conventional out-of-the-box sampling methods and the state-of-the-art logit-manipulation technique known as contrastive sampling. The key strength of STEER lies in its ability to significantly enhance the diversity of generated text (quantified by the number of normalised $n$-grams and the diversity score) without compromising coherence (as assessed by MAUVE, adversarial AUROC, and cosine similarity). Interestingly, contrastive sampling was found to underperform other methods, a phenomenon that persisted despite extensive tuning of the penalty term $\alpha$, which controls the strength of the degeneration penalty (see \citet{su2022contrastive}).

\begin{table}
\centering
\scriptsize
\setlength{\abovecaptionskip}{10pt}
\begin{tabular}{lccccc}
\toprule
& STEER & Greedy & Nucleus & Contrastive & \textit{Real} \\
\midrule
\textit{Jigsaw} & \textbf{0.94} $\pm$ 0.02 & 0.91 $\pm$ 0.03 & 0.90 $\pm$ 0.02 & 0.89 $\pm$ 0.01 & \textit{0.98} \\
\textit{QA} & \textbf{0.41} $\pm$ 0.03 & 0.35 $\pm$ 0.04 & 0.40 $\pm$ 0.03 & 0.29 $\pm$ 0.02 & \textit{0.55} \\
\bottomrule
\end{tabular}
\caption{Downstream performance comparison for Falcon-7B across two datasets: Jigsaw Toxic Comments and CommonsenseQA. Models were evaluated on five different splits of the real data.}
\label{tab:downstream_performance}
\vspace{-10pt}
\end{table}

An evaluation of the data quality produced by STEER was also conducted by training models on downstream tasks derived from the synthetic data, specifically on the \textit{Jigsaw} and \textit{CommonsenseQA} datasets. Both tasks involved classifiers and multiple-choice question-answering models respectively trained on synthetic data and evaluated on real data. Although the improvements were marginal, STEER consistently outperformed other methods (Table \ref{tab:downstream_performance}).

We also engaged expert evaluators to assess human preferences on generated \textit{Arxiv Hypotheses}. Although STEER achieved a win rate above 0.5 for low values of $\gamma$ and $\eta$ compared to the nucleus sampling method, this advantage diminished rapidly as both parameters exceeded 0.4 (Figure \ref{fig:combined_heatmap}). Along with the performance curves shown in Figure \ref{fig:vary_one}, these results hint at a narrow optimal range for hyperparameters, highlighting the need for specialised tuning techniques, including grid search on logarithmic scales \cite{goodfellow2016deep}. This figure also appears to show that good performance can be achieved with only 5-10 negative prompts. Further, MAUVE, measuring fidelity, declines as the number of negative samples increases. We hypothesise that this is because the conditional distribution becomes too constrained to produce coherent text.

Interestingly, the same blind evaluation with GPT-4 revealed more consistent win rates of STEER over nucleus, appearing to have a weaker gradient with respect to the magnitude of $\gamma$ and being much more sensitive to $\eta$ (Figure \ref{fig:combined_heatmap}). This observation, coupled with the lower STEER preferences among domain experts compared with general annotators, raises the conjecture that STEER may produce synthetically diverse and coherent examples that, however, may lack depth in semantic structure and plausibility. Further assessments across a wider task spectrum and more refined evaluation metrics will be crucial to validate or refute this hypothesis. For instance, filtering poor hypotheses from the initial fine-tuning dataset or performing additional quality ranking with fine-tuned models might align hypothesis generation better with domain experts.

\subsection{Limitations and future directions}
In this study, several limitations have been identified that must be acknowledged. One significant concern is the possible superficial optimisation of the evaluation metrics employed. The methods applied to gauge performance may not encompass the full complexity of the underlying processes. Furthermore, human preference evaluations have exhibited a lower alignment with STEER among astronomy domain experts for the hypothesis generation, pointing towards a potential disconnect between machine-optimised objectives and human-centric goals. Quantitative results on this discrepancy are presented in the appendix. The lack of variability across cosine similarity as a measure in our evaluation also presents a challenge, as it may not adequately capture the semantic nuances within the text. Moreover, the effectiveness of the negative prompting component of STEER is limited by the context window of the model, which constrains the number of negative prompts that can be used.

STEER is also twice as expensive at inference time as typical decoding methods due to the two forward passes: one through the base model $\mathrm{P}_\phi$ and one through the fine-tuned model $\mathrm{P}_\theta$, at each autoregressive step. As such, STEER may be more useful in low-data regimes where the issue is not with compute but rather the lack of data to learn from: STEER emphasises quality over speed.

Looking forward, several promising avenues can be explored to build upon this work. Investigating the effect of STEER on smaller models, such as GPT-2, might reveal insights into how logit-manipulation techniques can leverage the performance of smaller models to match or even surpass their larger counterparts. Implementing STEER in the latent space of hidden-layer representations rather than the token space could also provide a more nuanced control over text generation. Wandering off the main path to explore creative ideas and implementing a chain of thought in the form of back-and-forth checking could supplement this. Further, the application of STEER to bigger models warrants investigation, providing a comprehensive understanding of the method's scalability. 

Contrastive expert guidance functions as a regularisation technique during generation, compelling the model to create outputs that are more characteristic of the target domain. This approach can be viewed as an efficient method for enhancing fidelity, especially in scenarios where data is sparse. A valuable extension to this work would be a comparative analysis between the performance of contrastive expert guidance and the scaling of performance as the size of the real dataset available for fine-tuning increases. Such an analysis would shed light on whether contrastive guidance can serve as an advanced and efficient form of knowledge distillation, transcending the need for additional data collection and curation.

\section{Conclusion}
In this work, we illustrate that challenges with data coherency and diversity in Large Language Models (LLMs) can be mitigated through strategic logit reshaping during inference. Our novel method, STEER, leverages contrastive expert guidance and negative prompts from real and synthetic examples to balance adherence to data distribution and diversity. With its dynamic adjustments in the logit space, STEER outperforms previous approaches on the distinct tasks of hypothesis, toxic comment and commonsense question generation, as per conventional synthetic data generation metrics. 

Notably, STEER also exhibits a superior to capture semantic information in its synthetic examples that allow other models to achieve heightened downstream performance -- human preference, binary classification accuracy, and question-answering accuracy -- across all tasks. Additionally, our ablation studies underscore the control STEER provides over the coherency-diversity trade-off via its hyperparameters. This work not only addresses an existing gap in LLMs but also paves the path for more tailored applications of synthetic data generation in diverse domains.





\bibliography{references}  

\appendix

\section{Trade-off curve}
\begin{figure}[!h]
\centering
\includegraphics[width=0.85\columnwidth]{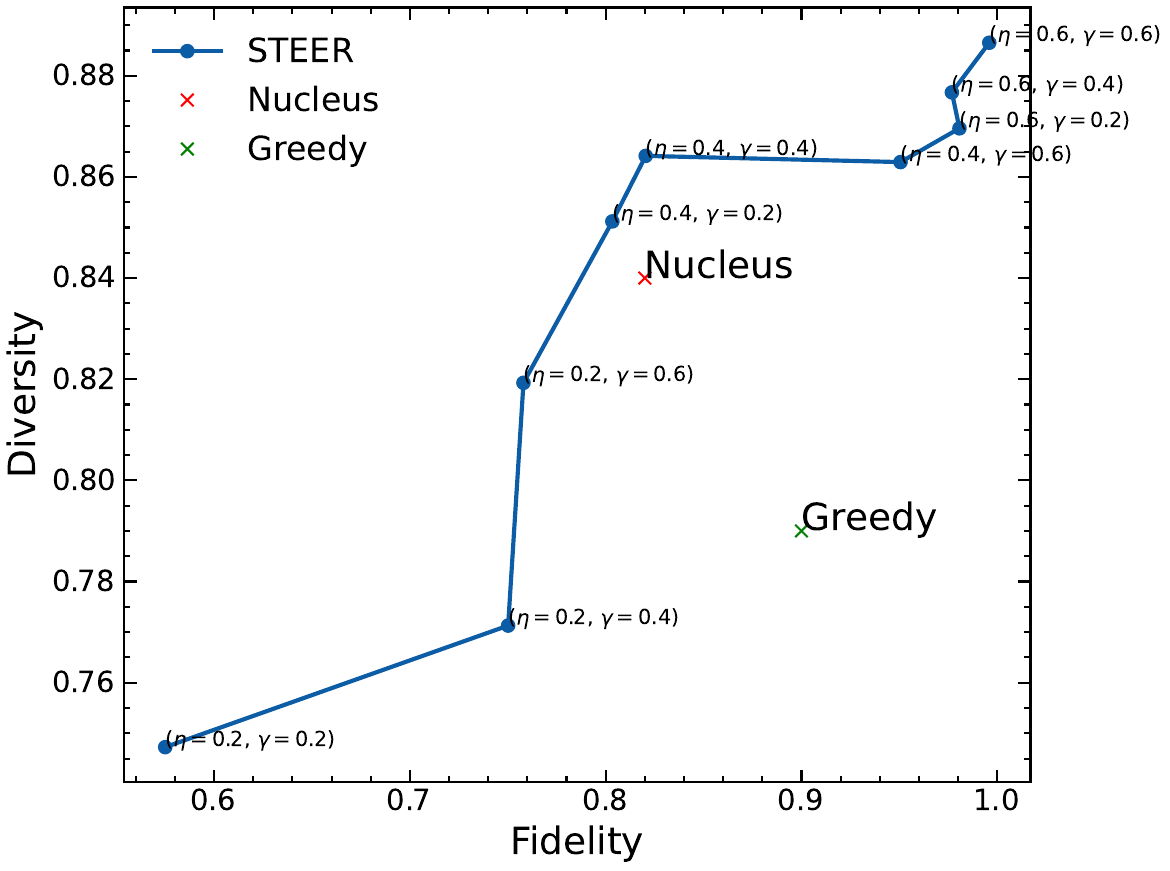}
\caption{Trade-off between MAUVE score and normalised $n$-grams score for 50 STEER generations in each hyperparameter combination.}
\label{fig:tradeoff}
\end{figure}

In Figure \ref{fig:tradeoff} we demonstrate how certain values of $\gamma$ (the contrastive expert guidance hyperparameter) and $\eta$ (the negative prompting hyperparameter) may be chosen to achieve Pareto-dominance of current sampling methods such as nucleus sampling and greedy decoding. That is, it is possible to choose $\eta$ and $\gamma$ so that the generated examples are better in all dimensions of evaluation (here, diversity and fidelity).

\section{Domain experts vs general evaluator}
\begin{figure}
\centering
\includegraphics[width=\columnwidth]{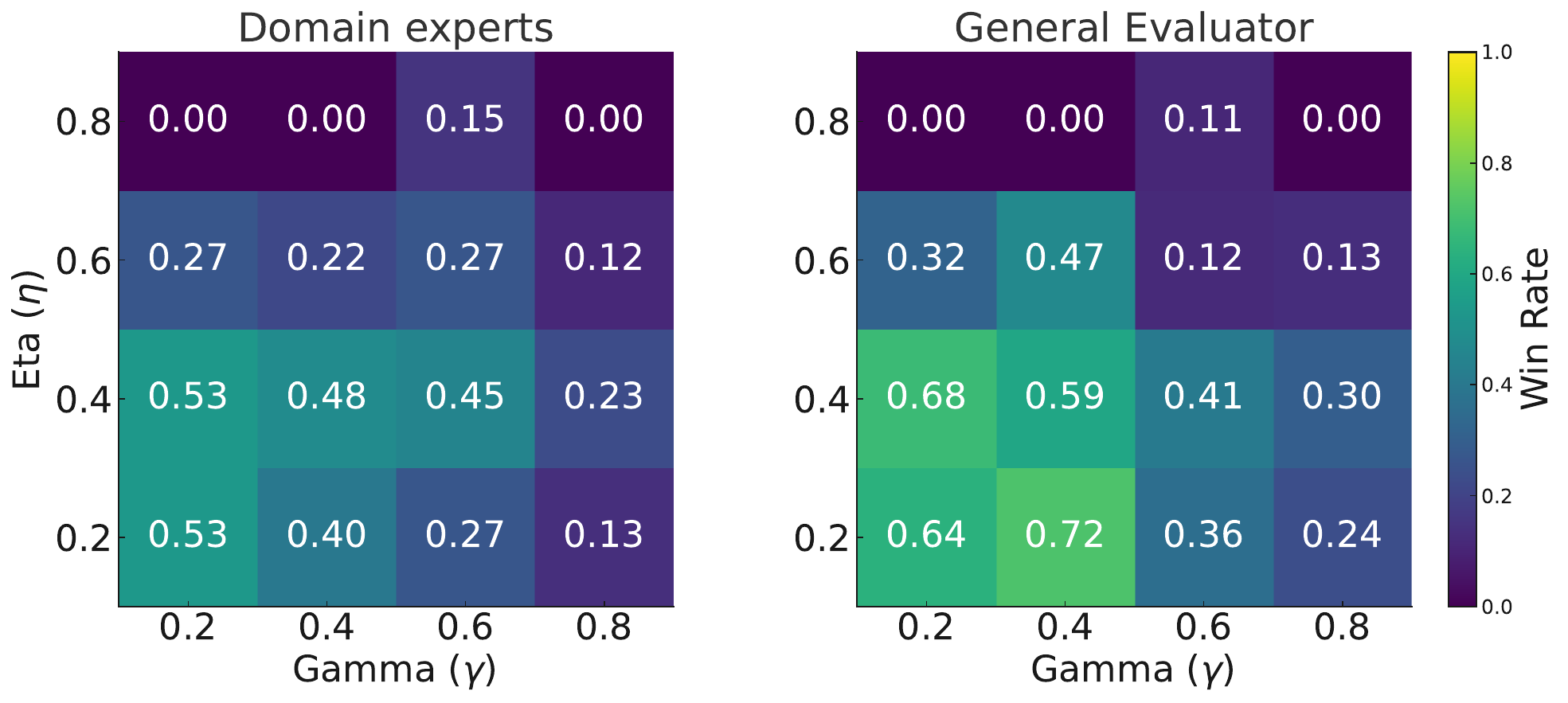}
\caption{Comparing the win rates of STEER vs nucleus for astronomy domain experts, defined by having postdoctoral qualifications in astronomy (three evaluators) compared with general annotators (five evaluators).}
\label{fig:domain_heatmap}
\end{figure}

In Figure \ref{fig:domain_heatmap} we show how the STEER generated hypotheses were significantly preferred by non-domain experts compared to domain experts in astronomy (as the generated hypotheses all concerned astronomy). Whilst both groups had a positive win-rate against nucleus sampling for small values of $\gamma$ and $\eta$, the general evaluators had much higher win rates and for larger values of these hyperparameters.

\section{UMAP embeddings to visualise dataset overlap}
\begin{figure*}
\centering
\includegraphics[width=0.65\textwidth]{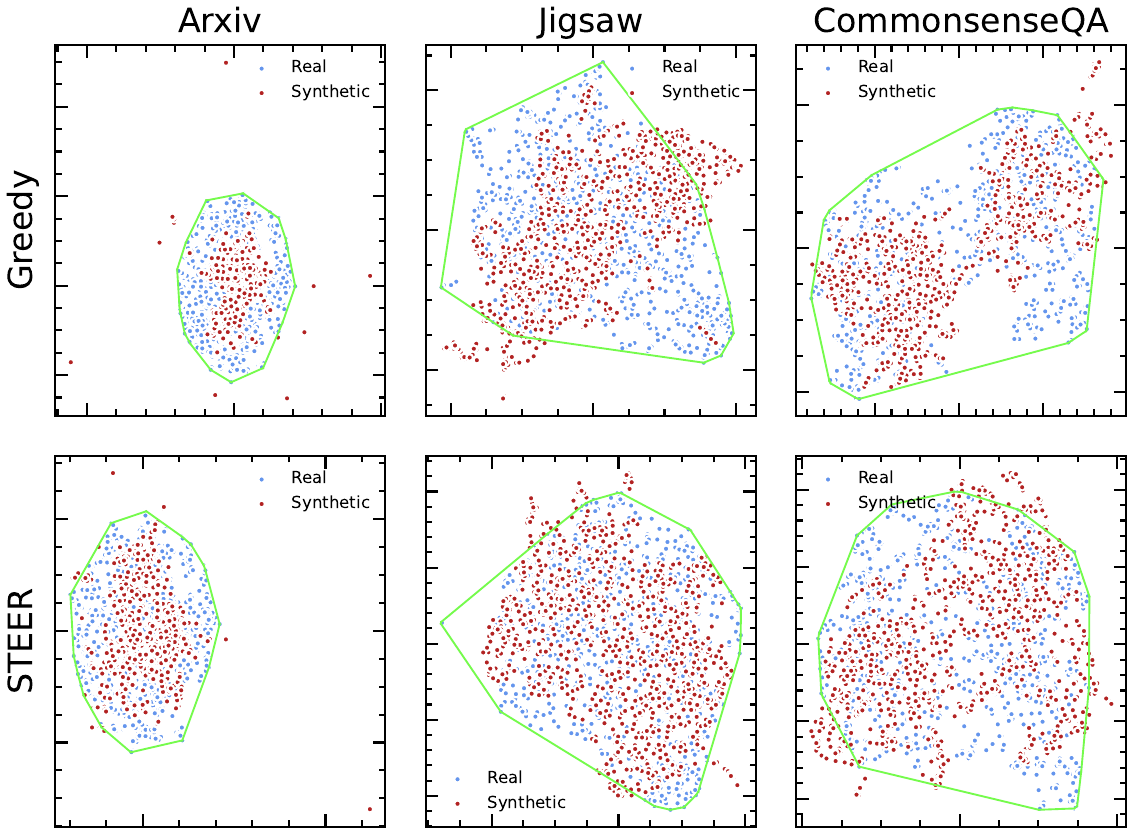}
\caption{UMAP visualisations of the embeddings for real and synthetic data across different models (Greedy, STEER) and datasets (Arxiv, Jigsaw, CommonsenseQA). Each subplot illustrates the 2D representation of embeddings, with the real embeddings colored in blue and the synthetic ones in red. The convex hull surrounding the real data is delineated by the green line. This visualisation provides a comparative insight into the geometric alignment and coverage of synthetic data with respect to real data across various contexts.}
\label{fig:umap}
\end{figure*}

To visualise the real and synthetic data distributions, we embedded the real and synthetic datasets using the OpenAI \texttt{text-embedding-ada-002} model and performed dimensionality reduction to two dimensions with UMAP. We then plotted both distributions on the same plot, as shown in Figure \ref{fig:umap}.

\label{app:umap}

\section{Convex hull precision and recall}
\label{app:convex}
It is possible to define precision and recall metrics based on the convex hull of the synthetic and real data distributions. Let $\mathcal{H}_r$ and $\mathcal{H}_s$ be the convex hulls defined by the real data points $\mathcal{D}_r=\{X_{r,i}\}^n_{i=1}$ and the synthetic data points $\mathcal{D}_s = \{X_{s, j}\}_{j=1}^m$, respectively.

The \textit{convex hull precision} is naturally defined as the proportion of synthetic data points that are accurately captured within $\mathcal{H}_r$. High precision means that the synthetic data points in $\mathcal{D}_s$ are mostly within the convex hull defined by the real data points in $\mathcal{D}_r$, indicating strong alignment with the real data distribution $\mathbb{P}_r$. Mathematically, this is:

$$\text{Convex Hull Precision} = \frac{|\{X_{s,j} \mid X_{s,j} \in \mathcal{H}_r\}_{j=1}^m|}{m}$$

\textit{Convex hull recall} reflects the proportion of real data points that are accurately captured within $\mathcal{H}_s$. High recall means that the real data points in $\mathcal{D}_r$ are mostly within the convex hull, indicating that the synthetic data in $\mathcal{D}_s$ has adequately covered the real data distribution $\mathbb{P}_r$. Formally:

$$\text{Convex Hull Recall} = \frac{|\{X_{r,i} \mid X_{r,i} \in \mathcal{H}_s\}_{i=1}^n|}{n}$$

These definitions provide a natural way to assess how well the synthetic dataset emulates the real dataset. High precision ensures that the synthetic data points do not stray far from the real data distribution, while high recall ensures that the synthetic data provides comprehensive coverage of the real data space. The F-score, the harmonic mean of precision and recall, is given by:

\[
\text{F-score} = \frac{2 \times P \times R}{P + R}
\]

Together, precision, recall, and the F-score give a good measure of the quality and representativeness of the synthetic data with respect to the real data. The precision and recall of all methods across the three datasets are shown in Table \ref{tab:convex_hull_metrics}. Interestingly, STEER had consistently high recall, suggesting its benefits lie largely in capturing a large range of real data points in the synthetic data distribution. Note that the convex hull was calculated with UMAP embeddings of 5 components. This was largely because the length of the embedding for each example was greater than the number of datapoints (1000), meaning the convex hull couldn't be easily calculated, and also due to computational concerns.

\begin{table}[htbp]
  \centering
  \scriptsize
  \setlength{\abovecaptionskip}{10pt}
  \begin{tabular}{l|c|c|c}
    \toprule
    & Convex Hull Precision & Convex Hull Recall & F-score \\
    \midrule
    \multicolumn{4}{l}{\textit{ArXiv Hypotheses}} \\
    Greedy & \textbf{0.997} & 0.949 & 0.972 \\
    Nucleus & 0.996 & 0.952 & 0.974 \\
    Contrastive & 0.996 & 0.867 & 0.927 \\
    \rowcolor{green!25} STEER & 0.994 & \textbf{0.963} & \textbf{0.978} \\
    \midrule
    \multicolumn{4}{l}{\textit{Jigsaw Toxic}} \\
    Greedy & 0.785 & 0.910 & 0.843 \\
    Nucleus & \textbf{0.802} & 0.807 & 0.805 \\
    Contrastive & 0.733 & 0.919 & 0.815 \\
    \rowcolor{green!25} STEER & 0.772 & \textbf{0.993} & \textbf{0.869} \\
    \midrule
    \multicolumn{4}{l}{\textit{CommonsenseQA}} \\
    Greedy & 0.886 & 0.969 & 0.926 \\
    \rowcolor{green!25} Nucleus & \textbf{0.945} & 0.953 & \textbf{0.949} \\
    Contrastive & 0.930 & 0.9610 & 0.945 \\
    STEER & 0.878 & \textbf{0.979} & 0.926 \\
    \bottomrule
  \end{tabular}
  \caption{Comparison of convex hull precision and convex hull recall for a fine-tuned Falcon-7B across three datasets: ArXiv Hypotheses, Jigsaw Toxic Comments, and CommonsenseQA.}
  \label{tab:convex_hull_metrics}
\end{table}


\section{Synthetic hypothesis examples}
\label{app:examples}

Included in Table \ref{tab:examples} are some examples of how STEER works as a combination of contrastive expert guidance and negative prompting. Whilst we normally supply full examples as negative prompts to STEER in the data generation process, these examples use a concept or keywords as a negative prompt to clearly illustrate the idea.

\newpage

\begin{table*}
\centering
\begin{tabular}{p{6.4cm}p{6.4cm}}
\toprule
\cellcolor{red!25}\textbf{\underline{Nucleus sampling}} & \cellcolor{green!25}\textbf{\underline{STEER}}
\\
\midrule
\rowcolor{blue!25} \multicolumn{2}{p{13.2cm}}{\textbf{\underline{Instruction:} \textit{``Generate a hypothesis about astronomy.''} \newline
\underline{Negative Prompt:} \textit{``Analogue satellite signatures''}}} \\
\midrule
Can the Gaia satellite detect the kinematic signature of a stellar warp in a star stream and establish whether or not it is co-located with a previously known warp?
&
Can convolutional neural networks be used to accurately deduce physical parameters and chemical abundances of stars from large spectroscopic and photometric data sets without applying any priors?
\\
\rowcolor{blue!25} \multicolumn{2}{p{13.2cm}}{\textbf{\underline{Instruction:} \textit{``Generate a hypothesis about astronomy.''} \newline
\underline{Negative Prompt:} \textit{``Gaia space mission''}}}\\
What is the best approximation of the local U-V velocity field due to orbital motion in the local area around the Sun, and how does it affect the calculation of the line-of-sight velocities of any given star?
&
Can a new method based on inferring the gravitational potential from the time-varying structure of a phase-space spiral in the Galactic potential accurately extract information and provide a complementary approach to traditional methods of dynamical mass measurements?
\\
\rowcolor{blue!25} \multicolumn{2}{p{13.2cm}}{\textbf{\underline{Instruction:} \textit{``Generate a hypothesis about astronomy.''} \newline
\underline{Negative Prompt:} \textit{``Gravitational potential analysis''}}}\\
What is the kinematics of the Galactic bar-bulge and how does it relate to the proper motions measured by the VVV and Gaia surveys?
&
Is the observed asymmetry of the tidal tails in certain star clusters the result of a stochastic process or due to additional dynamical processes beyond statistical evaporation?
\\
\rowcolor{blue!25} \multicolumn{2}{p{13.2cm}}{\textbf{\underline{Instruction:} \textit{``Generate a hypothesis about astronomy.''} \newline
\underline{Negative Prompt:} \textit{``General relativity''}}}\\
Does the 3D Two-Point Correlation Function (3D 2PCF) effectively quantify substructure in the Milky Way's stellar halo, and can it be used to constrain the accretion history of the galaxy?
&
Is the Galactic halo composed of two distinct parts, an inner and outer halo, and can these structures be identified through the use of chemical and kinematical analyses of open clusters?
\\
\rowcolor{blue!25} \multicolumn{2}{p{13.2cm}}{\textbf{\underline{Instruction:} \textit{``Generate a hypothesis about astronomy.''} \newline
\underline{Negative Prompt:} \textit{``Galactic halo''}}}\\
Can the analysis of LAMOST DR4 M giants plus Gaia DR2 TGAS parallaxes be used to construct a star catalogue with accuracies required for future spectroscopic surveys, while providing a homogeneous catalogue with uniform reduction and full parallax information?
&
Can the Gaia space mission's unique combination of accuracy, precision, and coverage allow for a thorough study of the structure and evolution of the Milky Way and lead to a better understanding of the galaxy's baryonic and dark matter components?
\\
\bottomrule \\
\end{tabular}
\caption{A comparative illustration of hypotheses generated using nucleus sampling and the STEER method with Falcon-7B. The underlined instruction and negative prompt guide the generation process. The left column, shaded in red, showcases a response utilising nucleus sampling. In contrast, the right column, shaded in green, presents a hypothesis generated by STEER.}
\label{tab:examples}
\end{table*}

\section{Formal STEER algorithm}
\label{app:algo}

\begin{algorithm}[H]
\SetAlgoLined
\KwResult{Synthetic dataset $\mathcal{D}_s$}
\textbf{Input:} Pre-trained transformer model $\mathrm{P}_\phi$, real dataset $\mathcal{D}_r$, hyperparameters $\gamma$ and $\eta$\\
Fine-tune $\mathrm{P}_\phi$ on $\mathcal{D}_r$ to obtain domain model $\mathrm{P}_\theta$\;
Initialise $\mathcal{D}_s = \emptyset$ (the synthetic dataset)\;
 \While{$\mathcal{D}_s$ not converged}{
  Initialise negative prompt set $\bar{c}$ from $\mathcal{D}_r$ and $\mathcal{D}_s$\;
  \For{each example $X_s$ to be generated}{
   Initialise $w_{<i}$ as the null token $\emptyset$\;
   \While{$w_{i}$ is not the end-of-sequence token}{
    Compute the logits of $w_{i}$ given $w_{j<i}$ using Equation \ref{eq:steer}\;
    Sample $w_{i}$ from the softmax distribution over logits\;
    Append $w_{i}$ to $w_{<i}$\;
   }
   Add generated example to $\mathcal{D}_s$\;
   }
 }
 \caption{STEER: Semantic Text Enhancement via Embedding Repositioning}
 \label{alg:steer}
\end{algorithm}


\end{document}